\documentclass{article}

\usepackage{epsfig}
\def\ArtWork#1{\noindent\hfill\epsfbox{#1}\hfill}

\begin{document}

\title{\textbf{Enhancing Support for Knowledge Works:} \\\emph{A relatively unexplored vista of computing research}\footnote{An abridged version of this article has been published as: A. Laha, G. Galarza and V. Agrawal, ``Support Systems for Knowledge Works: A perspective of future of Knowledge Management Service", \emph{2nd International Conference on Services in Emerging Markets (ICSEM 2011)}, 29th Sept- 01Oct,  Mumbai, 2011.}}
\author{Arijit Laha}

\date{}
\maketitle

%

\section{SSKW: A scenario}
Let us envision a new class of IT systems, the ``Support Systems for Knowledge Works" or SSKW. An SSKW can be defined as \emph{a system built for providing comprehensive support to human knowledge-workers while performing instances of complex knowledge-works of a particular type within a particular domain of professional activities.} To get an idea what an SSKW-enabled work environment can be like, let us look into a hypothetical scenario that depicts the interaction between a physician and a \emph{patient-care SSKW} during the activity of diagnosing a patient.

The patient-care task is practiced by health-care professionals, typically within organizational setups like hospitals. An instance of the task, known as a case, is carried out by a group of professionals (physicians, surgeons, nurses, laboratory technicians etc.) led by a physician (often known as the lead physician for the case) with the primary goal of restoring an ailing patient to state of health. However, the performance also serves various secondary goals achieved through capture and reuse of information about the case. The overall task is usually divided into subtasks or activities such as examination, identification of possible diseases, clinical tests, diagnosis, treatment, follow-up etc. The actions taken during these activities and their results have complex interrelationships. The patient-care SSKW \emph{realizes an integrated IT-based system platform which supports all the constituent activities in ways consistent with their interrelationships}.

Our hypothetical scenario depicts a particular activity by the lead physician (shall be referred as LP hereafter), i.e., diagnosing a patient $\cal P$ with the help of a patient-care SSKW. Making a diagnosis results in identifying a particular disease based on available evidence (e.g., symptoms, signs and medical history of the patient, results of various clinical tests conducted)  for which the patient will be treated. Such a scenario is described below.

For diagnosing $\cal P$, LP opens the case in SSKW and the following interactions take place:
\begin{enumerate}
  \item SSKW presents LP with a overview of the case depicting various activities involved, their dependencies and indicates that all pre-requisites for performing diagnosis are fulfilled and LP can start diagnosing; \label{l:overview}
  \item LP informs SSKW that she is staring diagnosis; \label{l:start}
  \item SSKW presents LP with the information required for diagnosis -- information from earlier activities of examination of the patient, possibilities considered, related tests (for confirming and/or eliminating possibilities) and their results; \label{l:present-info}
  \item SSKW also presents LP with a list of possible diseases based on the available evidence (i.e., results of examination and tests);\label{l:poss-dis}
  \item LP chooses a disease $D_1$ from the list as the tentative diagnosis;\label{l:choose-d}
  \item SSKW informs LP that\label{l:info-d}
  \begin{enumerate}
    \item  $D_1$ is caused by the pathogen $x$ (along with a link to more information on $x$ and $D_1$);\label{ll:path-x}
    \item in this hospital many (say, $n$) past cases with similar evidences the diagnoses were $D_1$;\label{ll:stat-d}
    \item however, in a significant (say, $p\%$) number of these cases the diagnosis was found wrong later -- SSKW provides a link to details;\label{ll:excep-d}
    \item also of these cases where wrong diagnoses were made, they are mostly re-diagnosed as diseases $D_2$ and $D_3$, which are caused by pathogens different but from same family as $x$ -- SSKW provides a link to a frequently referred analysis of these cases made earlier by another staff physician (also the link to the detailed analysis);\label{ll:analysis-d}
  \end{enumerate}
  \item SSKW also lets LP know that for this location (i.e., where the hospital is) $D_1$ is commonly occurs in a different time of year, so LP might like to check the following:\label{l:loc-time-d}
      \begin{enumerate}
        \item whether $\cal P$ traveled to some places where this pathogen is currently active?\label{ll:if-trav}
        \item $\cal P$ has contracted a mutated variant of $x$ -- SSKW presents relevant portions of a recently published research article where recent mutants of $x$ are reported;\label{ll:mutant}
      \end{enumerate}
  \item LP updates the history of $\cal P$ collected earlier with the information that $\cal P$ did not travel -- the information she elicited during examination but ignored to record;\label{l:update-hist}
  \item LP asks SSKW to keep her updated on new findings on mutation of $x$ (i.e., SSKW should notify her automatically whenever new information are available from reliable sources);\label{l:update-mut}
  \item LP asks SSKW whether there is any definitive procedure/test for eliminating the possibilities of $D_2$, $D_3$ or $D_1$ due to a mutant $x$;\label{l:elim-test}
  \item SSKW informs LP that there is a test but it is not very reliable -- SSKW presents summary of an analysis made by a staff physician on this matter as well as a link to the details of the analysis;\label{l:test-unrel}
  \item SSKW also informs  LP that researchers are working on devising such tests -- SSKW presents excerpts from a few recent publications on the topic;\label{l:test-res}
  \item LP asks SSKW to keep her updated on this subject (i.e., SSKW should notify her automatically whenever any significant progress is made);\label{l:test-res-update}
  \item LP uses SSKW to pull up some cases where the initial diagnosis was $D_1$ but subsequently found otherwise and successfully treated;\label{l:pull-cases}
  \item SSKW presents several cases matching the criteria;\label{l:ret-cases}
  \item LP studies the cases and\label{l:study-cases}
  \begin{enumerate}
    \item selects from them information related to how and when re-diagnosis made;\label{ll:select-info-fr-cases}
    \item asks SSKW to group them in terms of the stages of treatment when re-diagnosis is made, medications, signs and symptoms observed at that point of time and also compute correlations of these factors;\label{ll:analyze-info}
    \item LP interactively helps SSKW to develop the computation protocol for fine-tuning and presentation of results;\label{ll:help-analysis}
    \item SSKW presents the results to LP and stores the details of the protocol/process used in computation and presentation for future use;\label{ll:analysis-done-templatized}
  \end{enumerate}
  \item LP recognizes a pattern emerging from the analysis which can be used for determining whether the disease is other than $D_1$ in a relatively early stage of the treatment;\label{l:recog-patt}
  \item LP tells SSKW that she wants to consult some specialists/experts;\label{l:consult}
  \item SSKW returns a ranked list of physicians in the hospital who has good record of treating patients with diseases caused by $x$ and its relatives and/or authority on such diseases;\label{l:exp-loc}
  \item LP uses SSKW to organize and send relevant information, including the pattern she has unearthed to several of these specialists;\label{l:ask-exp}
  \item LP receives their responses through SSKW;\label{l:resp-exp}
  \item LP finds that specialists largely agree with the pattern she detected, however, some of them has sent back some additional observations;\label{l:exp-inputs}
  \item LP refines her idea on how to differentiate between $D_1$ and others from the additional inputs, formulates a strategy for discriminating between $D_1$ and others;\label{l:strat-discrim}
  \item SSKW captures the strategy and the process its development;\label{l:capt-strat}
  \item LP uses SSKW to organize all the information collected into groups supporting and dismissing various options available, interprets and evaluates them;\label{l:put-tug}
  \item LP decides to start treating $\cal P$ for non-mutant $D_1$ but also to plan the treatment and follow-up in such a way that any indication towards otherwise can be detected at earliest following the strategy she has developed;\label{l:decide-diag}
  \item LP uses SSKW to build her argument supporting the decision;\label{l:reason-diag}
  \item LP uses SSKW to capture the all information, along with their contextual relationships, developed during  diagnosis so that it becomes available for reuse later;\label{l:capt-diag}
  \item SSKW detects that the activity diagnosis is over; \label{l:diag-over}
  \item SSKW informs LP that now she can proceed to perform next activity of developing a ``treatment plan".\label{l:ready-for-tp}
\end{enumerate}

Note that, above scenario is conceived as an illustration of various aspects of a knowledge-intensive activity in general. Thus, we have taken some liberties in introducing some parts of interaction which may not be very realistic if viewed strictly in context of patient-care task, as it is practiced. However, such interactions can have vital impacts in other types of knowledge-work.

It can be easily recognized that the requirements for such support is not limited to diagnosis or patient-care. Similar problem situations occur with almost all complex knowledge-works, e.g., product development, basic and applied research, strategic business planning, organization design, health-care, urban planning, policy making etc. Each instance of such a task is a complex web of activities like planning, research, analysis, design, decision-making, prediction etc. All of them can benefit significantly from availability of a support system with capabilities similar to those envisaged above. In that sense, the SSKW can be thought rather as a \emph{class of systems}, of which the above patient-care SSKW is an example. Curiously enough, while the computer science research community pursued in earnest much more ambitious goal of building computer-based systems which can rival and/or replace human actors, the problem of designing systems like SSKW, which supports or assists human actors, has received much less attention.

Only within last few years we have started seeing results of works which can be viewed as attempts to building systems for supporting knowledge-workers. Two such systems are the ASAP \cite{Glasner:06} and the CODEX \cite{Pike:07}. While the former is designed for assisting genome researchers the later supports geography researchers. There are some ongoing research projects whose goals include some aspects of supports envisaged in SSKW but in domain-specific ways. Such initiatives include NEPOMUK - The Social Semantic Desktop \footnote{http://nepomuk.semanticdesktop.org/}, X-Media (Large Scale Knowledge Sharing and Reuse across Media)\footnote{http://www.x-media-project.org/}, DARPA-sponsored CALO (Cognitive Assistant that Learns and Organizes)\footnote{http://caloproject.sri.com/} etc.

Recently US Department of Health and Human Services (HHS) announced research funding for the ``Strategic Health IT Advanced Research Projects (SHARP)"\footnote{http://www.hhs.gov/news/press/2009pres/12/20091218c.html}. Much of the objectives of these projects deal with issues which can be identified with a robust patient-cage SSKW. We are likely to see many more such initiatives in near future. Thus, it is a high time for an attempt to understand the general issues involved in designing and implementing an effective SSKW. In the next section we shall analyze the scenario described above in order to identify some of the underlying capabilities which are required for an SSKW to support such interactions.

\section{Capabilities of an SSKW}
Our hypothetical SSKW above does a number of things. How such a set of functionalities can be brought together? Let us try to identify some of the general capabilities underlying them.

Line \ref{l:overview} of the scenario demonstrates a core capability of SSKW for recognizing the general structure of the patient-care task in terms of activities at the level of granularity they are practiced. SSKW also exhibits awareness regarding the progress made so far and possible activities which can be undertaken at this time. In line \ref{l:present-info} SSKW shows another vital capability of quite deep ``context-awareness" by presenting selectively information which are typically considered most relevant for performing diagnosis. Throughout the scenario we can observe SSKW's awareness about the required level of granularity of activity and information as well as their contextuality, explicitly or implicitly. Also, this awareness is dynamically maintained by the system as can be seen by its presentation of relevant information on the disease $D_1$ in lines \ref{l:info-d} following the choice of $D_1$ in line \ref{l:choose-d}. Additionally, these capabilities can be exploited to create SSKW functionality for guiding a novice worker as well as for SSKW to to marshal at a time its resources focused at
supporting a particular unit of activity.

Lines \ref{l:info-d} reveal another set of capabilities. Line \ref{ll:path-x} provides the information that infection by pathogen $x$ is likely cause of $D_1$. However, it also provides a link to more information about $x$ and $D_1$, which can be very useful for the physician if she is not familiar enough with them. Such mechanism can go a long way in addressing the issue of diversity in expertise level within a given community of workers with respect to a particular task/activity (as type or instance). Line \ref{ll:stat-d} reveals another facet of SSKW where it can reason that presenting directly a large number of similar cases can be of little use to the physician. Instead, it presents some computed statistics for these cases. This also requires that the SSKW need to be able to deal with information both in human comprehensible form as well as machine computable form. Line \ref{ll:excep-d} demonstrate even more interesting capability of goal-awareness in multiple levels. It not only understands the context and goal of the current diagnosis activity, it also recognizes their association with the goal of the overall task of patient-care, i.e., to get the diagnosis correct so that it forms the basis of a successful treatment. Line \ref{ll:analysis-d} provide more valuable elaboration of the information provided in previous line backed by a suitable documentation.

Lines \ref{l:loc-time-d} demonstrates SSKW's awareness of spatial and temporal contexts and ability to compare them \emph{semantically} with the cases available in the archive. On detection of conflict, it suggests action for disambiguation in line \ref{ll:if-trav} (which is done by the physician in line \ref{l:update-hist}. In line \ref{ll:mutant} SSKW suggest more radical means of resolving the conflict. To support these lines of the scenario, especially the last, SSKW needs to be able to access information from varied external source (say, journal archives) and associate the contents with context of interaction.

Next, a very interesting thing occurs in line \ref{l:update-mut}. The physician tasks SSKW to provide her automatically with new information about the research on mutation of $x$ as and when they become available. Here the physician is trying to protect herself against the common curse of knowledge-workers, ``professional obsolescence". Similar effort in her part can be observed in line \ref{l:test-res-update} with respect to development of new tests/procedures related to $D_1$. SSKW facilitates the effort by being in lookout for such information and supplying the user with the advances in knowledge in those areas.

From line \ref{l:pull-cases} to line \ref{l:strat-discrim}, the physician opts to do something very important, but not a common occurrence in a diagnosis activity. She accesses several cases likely to contain useful information, studies them. She identifies some relevant information (line for analyzing computationally with the help of SSKW (line \ref{ll:analyze-info}). However, SSKW is not capable to carry out the computation to her full satisfaction. So, she helps SSKW to develop the computations in line \ref{ll:help-analysis}. In the end, SSKW deliver results as required and also does something very interesting in line \ref{ll:analysis-done-templatized}. It keeps the computation and presentation protocols stored as templates for future use. In this way SSKW adds to its repertoire a new resource, which it may use in future to perform computations in similar contexts, or offer the template for required customization to users needing similar computation. An important facet of SSKW is revealed here. There will be times when an SSKW may not be able to provide satisfactory support. In such cases the user may need to give SSKW more detailed directions to produce required results. However, an SSKW tries to learn from the episode so that it can facilitate better level of support in future. This can be called as evolution of SSKW as a system.

Next, (line \ref{l:recog-patt}) the physician uses her knowledge, experience and the result of the analysis to detect a pattern (humans are good at it) which might lead to a development of new process/method for distinguishing occurrence $D_1$ from others definitively. However, she wants to verify whether she is in right track. So she ask SSKW about the people among her colleagues who are likely to able to help. In response, in line \ref{l:exp-loc}, SSKW performs ``expertise profiling" of the medical staff in the hospital, identify and rank them based on the current context. She sends her observations to the experts, and uses their responses to refine the process/method she has devised (line \ref{l:strat-discrim}). Then, in line \ref{l:capt-strat} SSKW captures the strategy and details about its bases.

The actions above carry great import for an SSKW as well as its user community. By its very nature, an SSKW is a social tool where a common case repository is used as well as contributed to by the whole community. Further, on the top of the cases, various types of analysis, interpretations etc. of collection of cases create immense value addition. For example, in the described scenario, the physician has benefited from results of such efforts (lines \ref{ll:analysis-d} and \ref{l:test-unrel}) by her colleagues. Preceding actions can be seen as her contributions at such higher level to the community in form of a possible new stategy/method, which can be tried and tested by her colleagues and if found viable can become part of the standard procedures. Such user activities add immensely to the value delivered by an SSKW and leads the SSKW and its user community in a trajectory of co-evolution towards more sophisticated capabilities.

In lines from \ref{l:put-tug} to \ref{l:reason-diag} the physician uses all information gathered so far (and knowledge gained in the process) in order to achieve the overall goal of the activity, namely, diagnosing so that she can move to the next stage of the task. She examines, interprets, evaluates and reasons with the information. The information generated during such activities form major source of reusable information in SSKW. Such information, at suitable level of granularity and contextualization, is available in adequate detail from a knowledge-worker only if she is supported adequately to articulate them at point of time close to their conception. Context-awareness of an SSKW allows it to marshal required resources for easy articulation of new information immediately after it is conceived and establishing its contextual relationships with other entities within the work-context. Then the SSKW captures and integrates (line \ref{l:capt-diag}) the information with its archive so that they, along with their required contextual properties such as argumentative structures, provenance, lineage and impact, are available for reuse in future instances of task performance.

Finally, in line \ref{l:diag-over} the SSKW, from the updated context, reasons that the goal for the current activity is achieved. Subsequently, SSKW, from its awareness about the structure of the task, guides the physician (line \ref{l:ready-for-tp} to undertake the next granule of activity, namely, preparation of a ``treatment plan" to be followed while applying treatment to $\cal P$.

\section{The challenge of designing an of SSKW}
A knowledge-work is a ``purposive" activity undertaken by human agents in order to solve a problem. A substantial knowledge-works is complex and evolving in nature. They are often identified as ``unstructured problems" \cite{Markus:02}. In fact, many of them show characteristics of wicked problems \cite{Rittel:73} which can be extremely difficult to solve because of incomplete, contradictory, and changing requirements that are often difficult to recognize. They exhibit complex interdependencies so that the effort to solve one aspect of a problem may reveal or even create other problems. Even when a solution is worked out, implementation of the solution may lead to unexpected outcomes. Additionally, in real-world setup, the people performing them can not be expected to possess equal level of expertise \cite{Markus:02}. All these issues together make the problem of designing an SSKW, especially if we seek its sustained usability by a community of users, extremely difficult one.

An SSKW can be considered as a ``cognitive technology (CT)" \cite{Dror:08} that \emph{tries its best} to support or assist a user to ``offload", i.e., simplify and/or expand in scope, some of the cognitive activities so that she can free up her cognitive capacities for performing more complex cognitive activities. Clearly, the challenge for a designer of SSKW lies in \emph{making the best possible attempt by the SSKW good enough}. We propose the following as some of the guiding principles in such endeavors.

\subsection{Principle of Resilience and Flexibility}
Given the general nature of a knowledge-work, if we attempt to build a very sophisticated system, based on a one-time understanding, however thorough it may be, of requirements, it is very likely to prove brittle soon in face of unpredictable changes in work environment and practices. Thus, we need to accept on the outset that we are not going to be able to anticipate future requirements and most possibly even all the current requirements. Consequently, it not going to always be possible for an SSKW to provide desired level of support. However, the SSKW must have \emph{multilayered capabilities} so that if and when required a user can easily build \emph{new} higher level capabilities leveraging lower level ones. Such a situation can be found in the scenario described above in line \ref{ll:analyze-info}, where the SSKW fails to provide the desired result. However, it provides enough lower level support based on which the physician can develop (line \ref{ll:help-analysis}) means of obtaining desired information.

The multilayered capabilities of an SSKW, additionally serve another crucial purpose. An SSKW is initially built around a particular structure of the target task. Let us call it the ``baseline structure" or BS of the task. We can not assume that BS is the best possible structure for the task. In fact, Markus et al. \cite{Markus:02} opine that there is no ``best structure" for a knowledge-work. Thus, it is to be expected that some users' (especially those with high level of expertise) needs will not be satisfied by the support implemented around the BS. An SSKW must allow them, at least within instance-specific or episodic context, to modify support for existing activities and/or create support for new activities using the lower level capabilities. The interaction from line \ref{l:pull-cases} to line \ref{l:strat-discrim} can be considered as an example of creating a new activity within the context of the case of $\cal P$.

\subsection{Principle of Evolution and Adaption}
Due to (right kind of) evolution, a system gets more efficient in what it does while adaption allows it to cope with new/changed demands on it due to changes in the environment it operates in. An SSKW, is by its very nature a communal tool, which poses some difficulty in its design which we shall examine later. However, it also opens up a window of opportunity which can be exploited by an SSKW in order to evolve and adapt. Every instance of performance using SSKW creates new resources, nominally the information related to the instance (line \ref{l:capt-diag}) and additionally other valuable resources such as those created in lines \ref{ll:analysis-done-templatized} and \ref{l:strat-discrim}.

An SSKW should be designed to so that it actively attempts maximize the scopes of utilization of these new resources. To achieve this effectively, an SSKW must employ various means to associate the specific situation which required resource with general situation(s) where they may be useful. For example, the SSKW may present the computation template to future users whose work-contexts are sufficiently similar to the context of its original creation. Now, based on the the users' feedbacks, explicit and implicit, regarding usefulness of the template, the SSKW can fine-tune the general scope of its use. Similarly, the strategy devised in line \ref{l:strat-discrim} can be evaluated from the impacts of its use and if found successful can be presented in context of future cases involving $D_1$, in similar way as the resources developed in past are presented in the current case in lines \ref{ll:analysis-d} and \ref{l:test-unrel}.

Also, adaption to changed environment can be achieved, either automatically or in a human supervised way, through study of emerging patterns from the episodic information. For example, consider the situation that in a large number of episodes, the workers have deviated from the baseline structure in a certain way and the impact of such performances are more positive than those episodes perform according to the BS. Since, such deviations are usually made when the workers need to do something extraordinary in response to some extraordinary (with respect to BS) problem, the above observation makes a strong case that the BS should be modified (or standardized) to incorporate the deviation and suitable support be arranged for it. Investigation into such occurrences can, in fact, lead to deeper organizational learning, in addition to update and adaptation of the SSKW with respect to changed environment.

\subsection{Principle of individual and communal usability}
Ease of use of an SSKW both for an individual perspective as well as communal perspective demands consideration of several factors. For an individual, even one with adequate knowledge and experience, cognitive factors come into play. Our cognitive capability of processing information is capacity-limited. Exact nature of this limitation is not yet known. It is likely to be influenced by neurobiological factors as well as social, cultural and technological ones \cite{Donald:01}. However, in real-world this is reflected in our practice of decomposing an activity into smaller sub-activities till we reach a level where an activity is cognitively manageable. As we discussed in context of patient-care, each community of professional practice has some standard structures of their tasks which the practitioners are familiar with. An SSKW must incorporate these structures and provide \emph{focused support} at the granularity level of their elements. This is reflected in supports for activities such as examination, diagnosis etc. in our hypothetical patient-care scenario.

Most important components of the focused support at granular level are those for \emph{information creation} and \emph{information consumption}. It is essential for an SSKW that the supports for these parts of activities be commensurate with the granularity level as well as cognitive constraints of the workers. Creation/production of new information results from a user gaining new knowledge or insight relevant to the problem-at-hand. This occurs at level of cognitively manageable granules of activities. An SSKW must provide means to the user for easy, in-time articulation, organization and recording of contextualized information at these granular levels as part of the performance. Such support can be observed in lines \ref{l:put-tug} to \ref{l:capt-diag} of our example scenario.

Typically, a knowledge-worker has large and complex information needs. However, these needs do not arise all together, rather they arise selectively in context of granules of activities. An SSKW, while seeking, retrieving and presenting information to a user, should make effort to choose them on the basis of possible relevance to the current granule of activity (see line \ref{ll:mutant}, where relevant part from larger documents are presented to the user). Also, if there is substantial amount of information which may inundate the user's cognitive capabilities, the SSKW should seek to extract and present relevant facts in terms of aggregates and statistics. Our example SSKW does the same in lines \ref{ll:stat-d}, \ref{ll:excep-d} and \ref{ll:analysis-d} since there are many past cases where the disease $D_1$ is diagnosed.

When a system is to be used by a community or group of individuals, it is very unrealistic to expect that all of them possess same level of expertise and thus can use the system with equal effectiveness \cite{Markus:02}. Nevertheless, reality demands that all of them be able to perform with, if not excellence, at least adequate competency. This is only possible if relative non-experts can be supported to bridge their knowledge-deficiency with reasonable investments in time and effort. We believe this issue to be of cardinal importance in an SSKW, since the activities supported by it themselves are typically of considerable complexity. Thus, an SSKW must be designed not only to deal with instance-specific information, but also generic information which can help a user to understand and conceptualize about various aspects of their work. In line \ref{ll:path-x}, our example SSKW provide a link to detailed information about the pathogen and disease caused by it. If the physician is unfamiliar with them, she can follow the link and gather relevant knowledge.

\subsection{Principle of coherent extensibility}
Given the complexity of knowledge-works and needs for large volume of diverse information while performing them, we believe that an SSKW should be designed as an open-ended system. It should be able to extend or enhance its capabilities by properly harnessing the powers of systems external to it. Such systems may include information repositories, analytic tools or even other SSKW. In our example scenario, in line \ref{l:poss-dis} the list of possible diseases can be sourced from standard disease databases, in line \ref{ll:mutant}, the excerpts of a article might have been sourced from an article archive (e.g., PubMed\footnote{http://www.ncbi.nlm.nih.gov/pubmed/}) over Internet. Similarly, the activities in lines \ref{ll:analyze-info} and \ref{ll:help-analysis} can actually be leveraging capabilities of an external statistical data analysis tool. For an SSKW designer the challenge lies in how to translate the needs and their context expressed by the SSKW users through their own domain-vocabulary, into the languages understandable by external tools and vice versa.

\section{Technological feasibility of SSKW}
The level of sophistication envisaged in our example scenario is perhaps not technologically achievable immediately. However, we believe that innovative application of state-of-the-art in computer science/IT research and technology can take us significant distance in the desired direction. In the following we shall discuss some relevant ideas and possibilities. Some of them are currently being explored in context of KwSS \cite{Laha:10}.

\begin{figure}
\epsfxsize=1.0\hsize \ArtWork{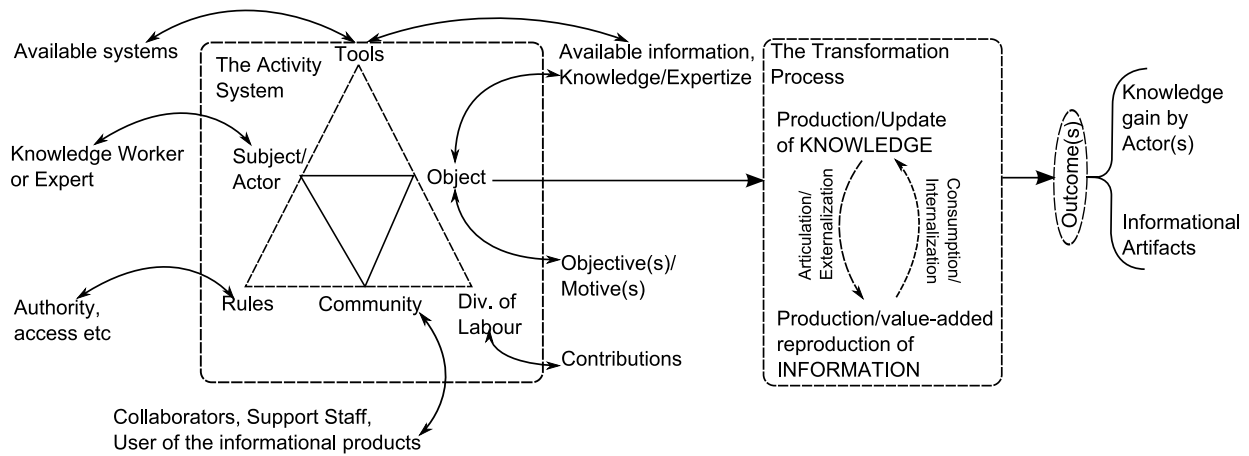} \caption{Activity theoretic view of knowledge-work.}\label{fig:kw-at}
\end{figure}


\subsection{Activity modeling}
An SSKW needs to to be aware of the structure of the tasks in terms of constituent activities and their interrelations (dependency, goal structure etc.) at the granularity level of their performance. Further, it also need to be aware of the context and semantics of activities as well as various resources (information, tools, computational protocols etc.) required for performing the activity. As means to capture them in order to form a nominal or baseline knowledge-base (KB) for SSKW, we need to create a rich and holistic yet formal representation of knowledge-intensive activities. Such a formalism may be co-opted from modern social sciences. For example, let us consider the model of a human activity as perceived in the ``Activity Theory (AT)" \cite{Kaptelinin:06}. Figure \ref{fig:kw-at} depicts the elements of the model and their correspondence to various entities involved in a knowledge-work.
According to AT, a human activity is primarily an interaction between an human actor/subject and an object, mediated by tools, in order to bring about changes in object. The social situated-ness of the activity is accounted in the model by the entity whose interactions with actor and object are mediated by rules and division of labor respectively. Here, object and tools are interpreted in broad sense. Object can be material as well as notional including intentions, goals etc. Similarly tools represents physical as well as mental/cognitive means which can be used to effect desired changes in object. These six entities together constitute an ``activity system" which makes possible to enact a ``transformation process" at the end of which the ``outcome" of the activity becomes available.

From SSKW design perspective, we can recognize various entities involved in a knowledge-work with respect to elements activity theoretic model of human activity as shown in figure \ref{fig:kw-at}. In an SSKW, we can model an activity by organizing information on related entities accordingly. In such a representation, the tools can encompass all the resources (information, systems, functionalities, expertise) and means to use them while the transformation process can be interpreted as the structure of the activity in terms of the sub-activities and their interrelationships. The sub-activities can also be represented in same way and analyzed recursively till we reach activity granules as they are practiced and/or are cognitively manageable. Use of such models are being explored in KwSS \cite{Laha:10}.

\subsection{Activity models in an SSKW}
Let us see what can be achieved by such models in an SSKW. A model of the target task type, populated with \emph{typological information} can serve as the definitional artifact of the system. The typological model (TM) is essentially a reference/nominal or baseline structure of the target which can be followed by a user of SSKW. On the other hand, SSKW can use the TM for tracking the activities as performed by a user. This enable SSKW to marshal resources for providing focused, context-aware support for the granular activities as well as detecting and recording deviations, if any, with respect to the TM made by a user.

An element of a TM can be designed to provide easy access to very rich resources by including among its contents detailed semantics (in form of domain ontology, glossary, thesauri etc.) and other relevant information (e.g., policy, legal restrictions) about the entity represented. This allow a user to access information for on-job-learning as well as the SSKW to perform sophisticated reasoning and inferencing.

Each performance of the supported task, defined by TM, is an episodic entity. It is unique with respect to other episodes in terms of its instance-specific information contents and sometimes structure also. Let us call the organization of instance-specific information as an ``episodic model (EM)". The TM in SSKW can serve as a template for initializing an episodic model. This has the advantage that each element of EM is associated with its typological counterpart so that  both a user and the system can easily access relevant typological information. A user may modify EM in order to perform optional and/or novel activities, as we have observed in our example scenario. Since an EM is a distinct instance-specific entity, the impacts of such modifications remain restricted within the particular episode.

\subsection{More of an SSKW}
While activity models form the foundation of an SSKW, it needs more capabilities. Sophisticated information retrieval (IR) is one of them. In an SSKW the information retrieval should be in context-aware manner. This can be formulated as a ``Case-based Reasoning (CBR)" problem, where the contextual information forms the attributes for computing similarity. It can be further refined by including in the attribute set, along with elements of episodic context also typological and conceptual context drawn from the typological model and knowledge-bases, such as domain ontologies associated with relevant elements of typological model. Such approach of computation results in supports exemplified by line \ref{ll:stat-d} in our example scenario.

However, the support in the next line (line \ref{ll:excep-d}), the SSKW requires to look deeper, i.e., beyond the immediate context, into the past cases to determine how did the acts of making the particular diagnosis impacted the higher level goal of treating a patient successfully. This demands that the standard CBR approach be extended to deal with the notion of desired goals and possible achievement categories (succeeded or failed) of other activities dependent on the current activity.

In the next line (line \ref{ll:analysis-d}) the SSKW provides more specific information about alternative diagnoses. Clearly, it uses the information developed for the previous line. But how did the SSKW decide that the result this particular resource, i.e., the computation protocol, will be useful to the physician? Well, the SSKW knows that in a past episode one of her colleagues has created an information resource, i.e., the analysis, and in the process she created and used this computation protocol. To deduce that the same may be useful in the current case requires in part of the SSKW to identify ``similarity of intentions" between the current physician and of her colleague when she performed the analysis.

Yes, both the activities have in common a same set of diseases to consider. But can the SSKW generalize the scope of utilization of the computation protocol further? The protocol is also useful in situations where there is a different but equally confusing set of alternative diagnoses. To detect the opportunity of using it in such situations, the SSKW requires to understand higher level intents involving deeper semantics. We believe that such generalizations of contexts/situations based on ``intent modeling" is going to play a vital role in SSKW of future. In fact, this capability of an SSKW greatly impacts its ease of evolution an adaption. We shall discuss more on them later.

An SSKW deals with resources in multiple forms, encompassing information in natural language text, logical and/or algorithmic entities, numerical data as result of measurements and computations, even possibly multimedia data. However, bulk of the information is contributed by human knowledge-workers in course of their using the system while performing their work. Also, whatever information is delivered by the system is meant for human comprehension. Thus, an SSKW must possess some capabilities for extracting logical facts and relationships from natural language texts which can be used by various computation-intensive components. It also should possess some capabilities of synthesizing natural language text encapsulating computational results. Compared with general setups for these tasks, an SSKW provides a much more conducive environment for implementing these capabilities. Such elements of an SSKW environment include well-defined granularity of texts, their detailed context and access to rich semantics associated with them through various activity models.

\subsection{Improving an SSKW}
Conceptually, an SSKW can be viewed as a \emph{learning (multi-)agent} \cite{Russell:10} but for a few crucial differences. Most significant among them arises due to accumulation of new resources. For an agent usually the challenge is to find better ways of serving the user with a well-defined and largely static repertoire of resources. In contrast, in an SSKW the volume of potential resources are ever-increasing. So, it needs to have additional capabilities for figuring out the usefulness of new resources. This can be very challenging to achieve. However, in case of SSKW we have the advantage that it is a system for assisting users rather than for replacing them. Consequently, we propose a pragmatic approach where systemic learning by SSKW can be complemented by suitable assistance from users.

An SSKW improves through evolution and adaption. Key to them is an SSKW's ability to identify and acquire resources and to extend their scopes of utilization. As mentioned earlier, new resources are created whenever a knowledge-worker performs using the SSKW. These resources have their immediate utilization within the particular episodic context. However, they contribute to the overall value of the SSKW only when they can be reused, directly or indirectly. Among such resources, information artifacts are reused by IR processes as described earlier. However, how can an SSKW figure out scopes of reuse for computational artifacts as developed in line \ref{ll:analysis-done-templatized} or procedural artifacts such as the one developed in line \ref{l:strat-discrim}? These can turn out to be highly complex problems. We believe a whole cornucopia of machine learning and data/information/relationship mining techniques can find their roles in solving them. Also, as suggested earlier, these problem can be made more tractable by utilizing suitable hints/clues or specifications from expert users.

For example, consider the computation template acquired by the SSKW in line \ref{ll:analysis-done-templatized}. This happened within a deep episodic context, which is unlikely to match another episodic context exactly. So the SSKW need to extract general aspects of the context within which the template may be reused. This problem might be formulated as a variation of ``reinforcement learning". The learning strategy can be that the SSKW, in an (future) episode, discovers that the context/situation has good match if some aspects of original contextual conditions of the template are relaxed. It presents the template to the worker, who may find it useful and uses it, find it irrelevant and rejects it or modify/customize it to suit better to her need and uses it. Based on such feedbacks over a number of such episodes, the SSKW can determine a sufficiently generalized context in which the template has good possibility of being reused. Also, at any point during this, a trusted human user can hugely expedite the process by identifying for the artifact a generalized context of utility.

\section{Conclusion}
We believe that SSKW, as a class of systems, going to play a big role in coming days. These systems can appear in various forms spanning from those used to provide personalized assistance to individuals to those used by large distributed teams for collaborating and orchestrating complex streams of information. However, realizing their potential, even at the level depicted in our example scenario, will require a multifaceted and concerted research effort encompassing a number of disciplines. Nevertheless, our experience with KwSS \cite{Laha:10} makes us believe that even the level of capabilities of an SSKW, which may be achieved with innovative applications of available technologies, can be of significant value.

\bibliographystyle{plain}
\bibliography{kwss}

\end{document}